\RequirePackage{fix-cm}
\documentclass[twocolumn]{svjour3}          
\smartqed  
\usepackage{graphicx}
\usepackage{epsfig}
\usepackage{amsmath}
\usepackage{amssymb}

\usepackage{xcolor}
\usepackage{soul}
\usepackage[utf8]{inputenc}
\usepackage[pagebackref=true,breaklinks=true,letterpaper=true,colorlinks,bookmarks=false,citecolor=cyan]{hyperref}

\usepackage[figuresright]{rotating}
\usepackage{textcomp,subfigure,algorithm,algorithmic,multirow,upgreek,tabularx}
\usepackage{graphics,gensymb}
\usepackage{bm,cases,threeparttable}

\usepackage{setspace}

\graphicspath{{Figures/}}
\usepackage{booktabs}

\usepackage{amsfonts}
 
\usepackage{caption}
\newcommand{\hao}[1]{\textcolor{black}{#1}}

\usepackage{enumitem}
\setitemize{noitemsep,topsep=0pt,parsep=0pt,partopsep=0pt}

\begin{document}

\title{Bipartite Graph Reasoning GANs for Person Pose and Facial Image Synthesis
}


\author{Hao Tang         \and
        Ling Shao \and
        Philip H.S. Torr \and 
        Nicu Sebe
}


\institute{Hao Tang (corresponding author) \at Department of Information Technology and Electrical Engineering, ETH Zurich, Switzerland. \email{hao.tang@vision.ee.ethz.ch}\\
             Ling Shao \at Terminus AI Lab, Terminus Group, China. \\
			Philip H.S. Torr \at
			Department of Engineering Science, University of Oxford, UK. \\
	        Nicu Sebe \at Department of Information Engineering and Computer Science (DISI), University of Trento, Italy.     
}

\date{Received: date / Accepted: date}

\maketitle

\begin{abstract}
We present a novel bipartite graph reasoning Generative Adversarial Network (BiGraphGAN) for two challenging tasks: person pose and facial image synthesis.
The proposed graph generator consists of two novel blocks that aim to model the pose-to-pose and pose-to-image relations, respectively. 
Specifically, the proposed bipartite graph reasoning (BGR) block aims to reason the long-range cross relations between the source and target pose in a bipartite graph, which mitigates some of the challenges caused by pose deformation.
Moreover, we propose a new interaction-and-aggregation (IA) block to effectively update and enhance the feature representation capability of both a person's shape and appearance in an interactive way.
To further capture the change in pose of each part more precisely, we propose a novel part-aware bipartite graph reasoning (PBGR) block to decompose the task of reasoning the global structure transformation with a bipartite graph into learning different local transformations for different semantic body/face parts. 
Experiments on two challenging generation tasks with three public datasets demonstrate the effectiveness of the proposed methods in terms of objective quantitative scores and subjective visual realness.
The source code and trained models are available at
\url{https://github.com/Ha0Tang/BiGraphGAN}.
\end{abstract}
\section{Introduction}
\label{sec:intro}

\sloppy

\begin{figure*}[!t]
	\centering
	\includegraphics[width=0.9\linewidth]{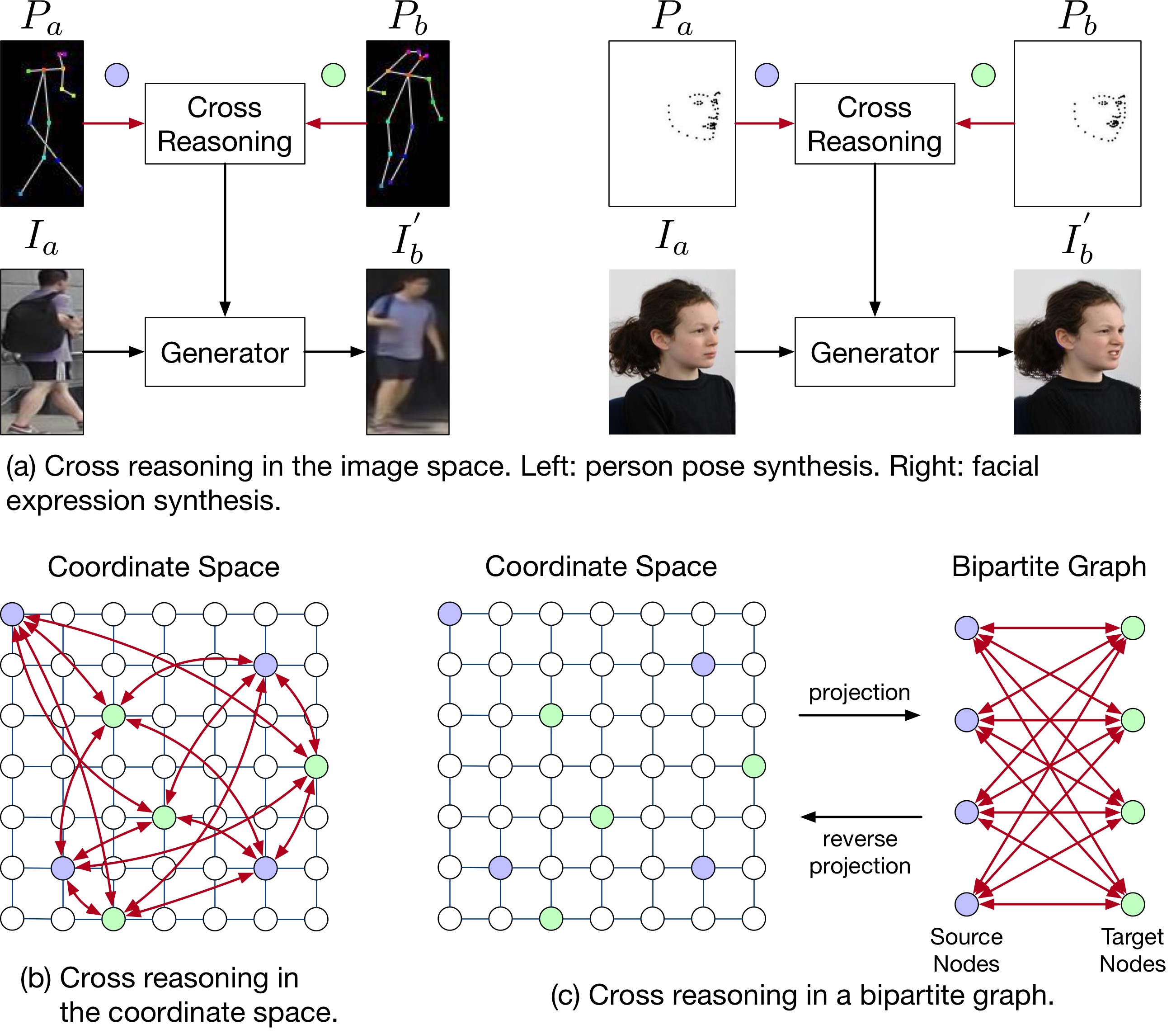}
	\caption{Illustration of our motivation. We propose a novel BiGraphGAN (Figure~(c)) to capture the long-range cross relations between the source pose $P_a$ and the target pose $P_b$ in a bipartite graph. The node features from both the source and target poses in the coordinate space are projected into the nodes in a bipartite graph, thereby forming a fully connected bipartite graph. After cross-reasoning the graph, the node features are projected back to the original coordinate space for further processing.}
	\label{fig:motivation}
\end{figure*}

In this paper, we focus on translating a person image from one pose to another and a facial image from one expression to another, as depicted in Figure~\ref{fig:motivation}(a).
Existing person pose and facial image generation methods, such as \cite{ma2017pose,ma2018disentangled,siarohin2018deformable,tang2019cycle,albahar2019guided,esser2018variational,zhu2019progressive,chan2019everybody,balakrishnan2018synthesizing,zanfir2018human,liang2019pcgan,liu2019liquid,tang2019cycle,zhang2020cross} typically rely on convolutional layers. 
However, due to the physical design of convolutional filters, convolutional operations can only model local relations.
To capture global relations, existing methods such as \cite{zhu2019progressive,tang2019cycle} inefficiently stack multiple convolutional layers to enlarge the receptive fields to cover all the body joints from both the source pose and the target pose.
However, none of the above-mentioned methods explicitly consider modeling the cross relations between the source and target pose.

\hao{Rather than relying solely on convolutions/Transformers in
the coordinate space to implicitly capture the cross relations between the source pose and the target pose, we propose to construct a latent interaction space where global or long-range (can also be understood as long-distance, which means that the distance between the same joint on the source pose and the target pose very long) reasoning can be performed directly. Within this interaction space, a pair of source and target joints that share similar semantics (e.g., the source left-hand and the target left-hand joints) are represented by a single mapping, instead of a set of scattered coordinate-specific mappings. Reasoning the relations of multiple different human joints is thus simplified to modeling those between the corresponding mappings in the interaction space. We thus build a bipartite graph connecting these mappings within the interaction space and perform relation reasoning over the bipartite graph. After the reasoning, the updated information is then projected back to the original coordinate space for the generation task. Accordingly, we design a novel bipartite graph reasoning (BGR) to efficiently implement the coordinate-interaction space mapping process, as well as the
cross-relation reasoning by graph convolution network (GCNs).}

In this paper, we propose a novel bipartite graph reasoning Generative Adversarial Network (BiGraphGAN), which consists of two novel blocks, i.e., a bipartite graph reasoning (BGR) block and an interaction-and-aggregation (IA) block.
The BGR block aims to efficiently capture the long-range cross relations between the source pose and the target pose in a bipartite graph (see Figure~\ref{fig:motivation}(c)). 
Specifically, the BGR block first projects both the source pose  and target pose feature from the original coordinate space onto a bipartite graph.
Next, the two features are represented by a set of nodes to form a fully connected bipartite graph, on which long-range cross relation reasoning is performed by GCNs. 
To the best of our knowledge, we are the first to use GCNs to model the long-range cross relations for solving both the challenging person pose and facial image generation tasks.
After reasoning, we project the node features back to the original coordinate space for further processing. 
Moreover, to further capture the change in pose of each part more precisely, we further extend the BGR block to the part-aware bipartite graph reasoning (PBGR) block, which can capture the local transformations among body parts.

Meanwhile, the IA block is proposed to effectively and interactively enhance a person's shape and appearance features.
We also introduce an attention-based image fusion (AIF) module to selectively generate the final result using an attention network.
Qualitative and quantitative experiments on two challenging person pose generation datasets, i.e., Market-1501 \cite{zheng2015scalable} and DeepFashion \cite{liu2016deepfashion}, demonstrate that the proposed BiGraphGAN and BiGraphGAN++ generate better person images than several state-of-the-art methods, i.e., PG2~\cite{ma2017pose}, DPIG~\cite{ma2018disentangled}, Deform~\cite{siarohin2018deformable}, C2GAN~\cite{tang2019cycle}, BTF~\cite{albahar2019guided}, VUNet~\cite{esser2018variational}, PATN~\cite{zhu2019progressive}, PoseStylizer~\cite{huang2020generating}, and XingGAN \cite{tang2020xinggan}.

Lastly, to evaluate the versatility of the proposed BiGraphGAN, we also investigate the facial expression generation task on the Radboud Faces dataset~\cite{langner2010presentation}.
Extensive experiments show that the proposed method achieves better results than existing leading baselines, such as Pix2pix~\cite{isola2017image}, GPGAN~\cite{di2018gp}, PG2~\cite{ma2017pose}, CocosNet \cite{zhang2020cross}, and C2GAN \cite{tang2019cycle}.

The contributions of this paper are summarized as follows:
\begin{itemize}[leftmargin=*]
	\item We propose a novel bipartite graph reasoning GAN (BiGraphGAN) for person pose and facial image synthesis. The proposed BiGraphGAN aims to progressively reason the pose-to-pose and pose-to-image relations via two novel blocks.
	\item We propose a novel bipartite graph reasoning (BGR) block to effectively reason the long-range cross relations between the source and target pose in a bipartite graph, using GCNs.
	\item We introduce a new interaction-and-aggregation (IA) block to interactively enhance both a person's appearance and shape feature representations.
	\item We decompose the process of reasoning the global structure transformation with a bipartite graph into learning different local transformations for different semantic body/face parts, which captures the change in pose of each part more precisely. 
	To this end, we propose a novel part-aware bipartite graph reasoning (PBGR) block to capture the local transformations among body parts.
	\item Extensive experiments on both the challenging person pose generation and facial expression generation tasks with three public datasets demonstrate the effectiveness of the proposed method and its significantly better performance compared with state-of-the-art methods.
\end{itemize}

Some of the material presented here appeared in \cite{tang2020bipartite}. The current paper extends \cite{tang2020bipartite} in several ways:
\begin{itemize}[leftmargin=*]
\item More detailed analyses are presented in the ``Introduction'' and ``Related Work'' sections, which now include very recently published papers dealing with person pose and facial image synthesis.
\item We propose a novel PBGR block to capture the local transformations among body parts. Equipped with this new module, our BiGraphGAN proposed in \cite{tang2020bipartite} is upgraded to BiGraphGAN++. 
\item \hao{We present an in-depth description of the proposed method, providing the architectural and implementation details, with special emphasis on guaranteeing the reproducibility of our experiments. The source code is also available online.}
\item We extend the experimental evaluation provided in \cite{tang2020bipartite} in several directions. 
First, we conduct extensive experiments on two challenging tasks with three popular datasets, demonstrating the wide application scope of the proposed BiGraphGAN and BiGraphGAN++.
Second, we also include more state-of-the-art baselines (e.g., PoseStylizer~\cite{huang2020generating} and XingGAN \cite{tang2020xinggan}) for the person pose generation task, and observe that the proposed BiGraphGAN and BiGraphGAN++ achieve better results than both methods. 
Lastly, we conduct extensive experiments on the facial expression generation task, demonstrating both quantitatively and qualitatively that the proposed method achieves much better results than existing leading methods such as Pix2pix~\cite{isola2017image}, GPGAN~\cite{di2018gp}, PG2~\cite{ma2017pose}, CocosNet \cite{zhang2020cross}, and C2GAN \cite{tang2019cycle}.
\end{itemize}
\begin{figure*}[!t]
	\centering
	\includegraphics[width=1\linewidth]{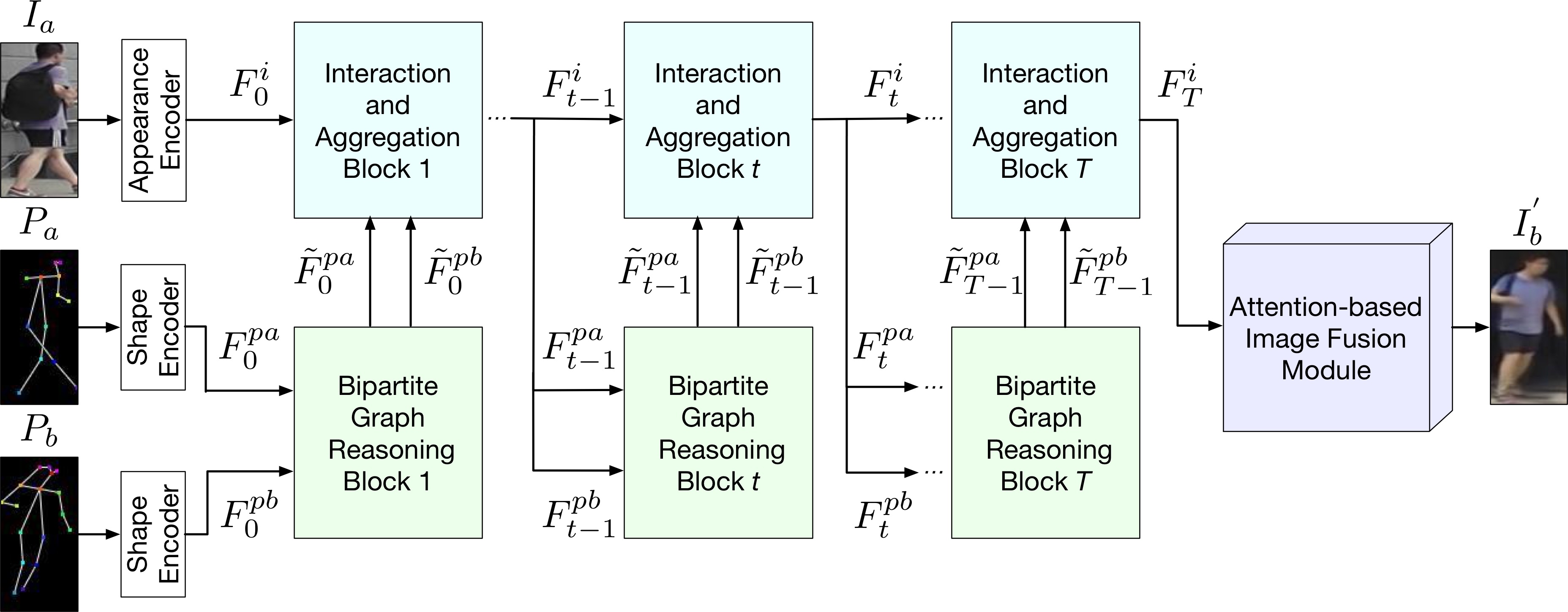}
	\caption{Overview of the proposed graph generator, which consists of a sequence of bipartite graph reasoning (BGR) blocks, a sequence of interaction-and-aggregation (IA) blocks, and an attention-based image fusion (AIF) module. The BGR blocks aim to reason the long-range cross relations between the source pose and the target pose in a bipartite graph. The IA blocks aim to interactively update a person's appearance and shape feature representations. The AIF module aims to selectively generate the final result via an attention network.
		The symbols $F^i{=}\{F^i_j\}_{j=0}^T$, $F^{pa}{=}\{F^{pa}_j\}_{j=0}^{T{-}1}$, $F^{pb}{=}\{F^{pb}_j\}_{j=0}^{T{-}1}$, $\tilde{F}^{pa}{=}\{\tilde{F}^{pa}_j\}_{j=0}^{T{-}1}$, and $\tilde{F}^{pb}{=}\{\tilde{F}^{pb}_j\}_{j=0}^{T{-}1}$ denote the appearance codes, the source shape codes, the target shape codes, the updated source shape codes, and the updated target shape codes, respectively.
	}
	\label{fig:method}
\end{figure*}

\section{Related Work}

\noindent \textbf{Generative Adversarial Networks (GANs)} \cite{goodfellow2014generative} have shown great potential in generating realistic images \cite{shaham2019singan,karras2019style,brock2019large,zhang2022unsupervised,zhang20213d,tang2021total,tang2020unified}.
For instance, Shaham et al. proposed an unconditional SinGAN~\cite{shaham2019singan} which can be learned from a single image.
Moreover, to generate user-defined images, several conditional GANs (CGANs) \cite{mirza2014conditional} have recently been proposed.
A CGAN always consists of a vanilla GAN and external guidance information such as class labels \cite{wu2019relgan,choi2018stargan,zhang2018sparsely,tang2019attribute}, text descriptions \cite{xu2022predict,tao2022df}, segmentation maps \cite{tang2019multi,park2019semantic,tang2020local,liu2020exocentric,wu2022cross,wu2022cross_tmm,tang2022local,ren2021cascaded,tang2021layout,tang2020dual}, attention maps \cite{kim2019u,tang2019attention,mejjati2018unsupervised,tang2021attentiongan}, or human skeletons \cite{albahar2019guided,balakrishnan2018synthesizing,zhu2019progressive,tang2018gesturegan,tang2020xinggan}.

In this work, we focus on the person pose and facial expression generation tasks, which aim to transfer a person image from one pose to another and a facial image from one expression to another, respectively.

\noindent \textbf{Person Pose Generation} is a challenging task due to the pose deformation between the source image and the target image.
Modeling the long-range relations between the source and target pose is the key to solving this.
However, existing methods, such as \cite{balakrishnan2018synthesizing,albahar2019guided,esser2018variational,chan2019everybody,zanfir2018human,liang2019pcgan,liu2019liquid}, are built by stacking several convolutional layers, which can only leverage the relations between the source pose and the target pose locally.
For instance, Zhu et al.~\cite{zhu2019progressive} proposed a pose-attentional transfer block (PATB), in which the source and target poses are simply concatenated and then fed into an encoder to capture their dependencies.

\noindent \textbf{Facial Expression Generation} aims to translate one facial expression to another \cite{tang2019expression,tang2019cycle,pumarola2020ganimation,choi2018stargan}.  
For instance, 
Choi et al. \cite{choi2018stargan} proposed a scalable method that can perform facial expression-to-expression translation for multiple domains using a single model.
Pumarola et al. \cite{pumarola2020ganimation} introduced a GAN conditioning scheme based on action unit (AU) annotations, which describes in a continuous manifold the anatomical facial movements defining a human expression.
Finally, Tang et al. \cite{tang2019cycle} proposed a novel Cycle in Cycle GAN (C2GAN) for generating human faces and bodies.

Unlike existing person pose and facial expression generation methods, which model the relations between the source and target poses in a localized manner, we show that the proposed BGR block can bring considerable performance improvements in the global view.

\noindent \textbf{Graph-Based Reasoning.} Graph-based approaches have been shown efficient at reasoning relations in many computer vision tasks such as semi-supervised classification \cite{kipf2017semi}, video recognition \cite{wang2018videos}, crowd counting~\cite{chen2020relevant}, action recognition \cite{yan2018spatial,peng2020mix}, face clustering \cite{wang2019linkage,yang2019learning}, and semantic segmentation \cite{chen2019graph,zhang2019dual}.

In contrast, to these graph-based reasoning methods, which model the long-range relations within the same feature map to incorporate global information, we focus on developing two novel BiGraphGAN and BiGraphGAN++ frameworks that reason and model the long-range cross relations between different features of the source and target pose in a bipartite graph.
Then, the cross relations are further used to guide the image generation process (see Figure~\ref{fig:motivation}). 
This idea has not been investigated in existing GAN-based person image generation or even image-to-image translation methods.

\begin{figure*}[!t]
	\centering
	\includegraphics[width=1\linewidth]{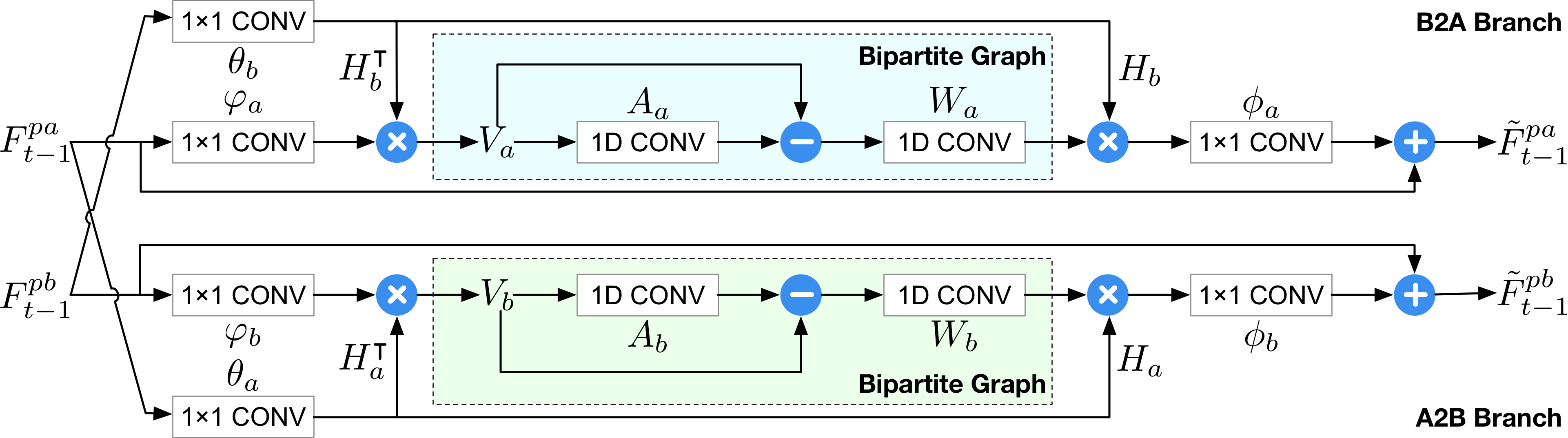}
	\caption{Illustration of the proposed bipartite graph reasoning (BGR) block $t$, which consists of two branches, i.e., B2A and A2B. Each branch aims to model cross-contextual information between shape features $F_{t-1}^{pa}$ and $F_{t-1}^{pb}$ in a bipartite graph via GCNs.}
	\label{fig:blocks}
\end{figure*}

\section{Bipartite Graph Reasoning GANs}

We start by introducing the details of the proposed bipartite graph reasoning GAN (BiGraphGAN), which consists of a graph generator $G$ and two discriminators (i.e., the appearance discriminator $D_a$ and shape discriminator $D_s$).
An illustration of the proposed graph generator $G$ is shown in Figure~\ref{fig:method}. It contains three main parts, i.e., a sequence of bipartite graph reasoning (BGR) blocks modeling the long-range cross relations between the source pose $P_a$ and the target pose $P_b$, a sequence of interaction-and-aggregation (IA) blocks interactively enhancing both the person's shape and appearance feature representations, and an attention-based image fusion (AIF) module attentively generating the final result~$I_b^{'}$.
In the following, we first present the proposed blocks and then introduce the optimization objective and implementation details of the proposed BiGraphGAN.

Figure~\ref{fig:method} shows the proposed graph generator $G$, whose inputs are the source image $I_a$, the source pose $P_a$, and the target pose $P_b$.
The generator $G$ aims to transfer the pose of the person in the source image $I_a$ from the source pose $P_a$ to the target pose $P_b$, generating the desired image $I_b^{'}$.
Firstly, \hao{$I_a$, $P_a$, and $P_b$ are separately fed into three encoders to obtain the initial appearance code $F_0^i$, the initial source shape code $F_0^{pa}$, and the initial target shape code $F_0^{pb}$.}
Note that we use the same shape encoder to learn both $P_a$ and $P_b$, i.e., the two shape encoders used for learning the two different poses share weights.

\subsection{Pose-to-Pose Bipartite Graph Reasoning}
The proposed BGR block aims to reason the long-range cross relations between the source pose and the target pose in a bipartite graph.
All BGR blocks have an identical structure, as illustrated in Figure~\ref{fig:method}. 
\hao{Consider the $t$-th block given in Figure~\ref{fig:blocks}, whose inputs are the source shape code $F_{t-1}^{pa}$ and the target shape code $F_{t-1}^{pb}$.}
The BGR block aims to reason these two codes in a bipartite graph via GCNs and outputs new shape codes.
It contains two symmetrical branches (i.e., the B2A branch and A2B branch) because a bipartite graph is bidirectional. As shown in Figure~\ref{fig:motivation}(c), each source node is connected to all the target nodes; at the same time, each  target node is connected to all the source nodes.
In the following, we describe the detailed modeling process of the B2A branch.
Note that the A2B branch is similar.

\noindent \textbf{From Coordinate Space to Bipartite-Graph Space.}
Firstly, we reduce the dimension of the source shape code $F_{t-1}^{pa}$ with the function $\varphi_a(F_{t-1}^{pa}) {\in} \mathbb{R}^{C \times D_a}$, where $C$ is the number of feature map channels, and $D_a$ is the number of nodes of $F_{t-1}^{pa}$.
Then we reduce the dimension of the target shape code $F_{t-1}^{pb}$ with the function $\theta_b(F_{t-1}^{pb}) {=} H_b^\intercal {\in} \mathbb{R}^{D_b  \times C}$, where $D_b$ is the number of nodes of $F_{t-1}^{pb}$.
Next, we project $F_{t-1}^{pa}$ to a new feature $V_a$ in a bipartite graph using the projection function $H_b^T$. 
Therefore we have:
\begin{equation}
\begin{aligned}
V_a = H_b^\intercal \varphi_a(F_{t-1}^{pa}) = \theta_b(F_{t-1}^{pb}) \varphi_a(F_{t-1}^{pa}),
\end{aligned}
\end{equation}
where both functions $\theta_b(\cdot)$ and $\varphi_a(\cdot)$ are implemented using a $1{\times1}$ convolutional layer. 
This results in a new feature $V_a {\in} \mathbb{R}^{D_b \times D_a}$ in the bipartite graph, which represents the cross relations between the nodes of the target pose $F_{t-1}^{pb}$ and the source pose $F_{t-1}^{pa}$ (see Figure~\ref{fig:motivation}(c)).

\noindent \textbf{Cross Reasoning with a Graph Convolution.}
After projection, we build a fully connected bipartite graph with adjacency matrix $A_a {\in} \mathbb{R}^{D_b \times D_b}$.
We then use a graph convolution to reason the long-range cross relations between the nodes from both the source and target poses, which can be formulated as:
\begin{equation}
\begin{aligned}
M_a = ({\rm I} - A_a) V_a W_a,
\end{aligned}
\end{equation}
where $W_a {\in} \mathbb{R}^{D_a \times D_a}$ denotes the trainable edge weights.
We follow \cite{chen2019graph,zhang2019dual} and use Laplacian smoothing  \cite{chen2019graph,li2018deeper} to propagate the node features over the bipartite graph. 
The identity matrix~${\rm I}$ can be viewed as a residual sum connection to alleviate optimization difficulties. 
Note that we randomly initialize both the adjacency matrix $A_a$ and the weights $W_a$, and then train them by gradient descent in an end-to-end manner.

\noindent \textbf{From Bipartite-Graph Space to Coordinate Space.}
After the cross-reasoning, the new updated feature $M_a$ is mapped back to the original coordinate space for further processing.
Next, we add the result to the original source shape code $F_{t-1}^{pa}$ to form a residual connection \cite{he2016deep}.
This process can be expressed as:
\begin{equation}
\begin{aligned}
\tilde{F}_{t-1}^{pa} = \phi_a(H_b M_a) + F_{t-1}^{pa},
\end{aligned}
\end{equation}
where we reuse the projection matrix $H_b$ and apply a linear projection
$\phi_a(\cdot)$ to project $M_a$ back to the original coordinate space. 
Therefore, we obtain the new source feature $\tilde{F}_{t-1}^{pa}$, which has the same dimension as the original one $F_{t-1}^{pa}$.

Similarly, the A2B branch outputs the new target shape feature $\tilde{F}_{t-1}^{pb}$. 
Note that the idea behind the proposed BGR block was inspired by the GloRe unit proposed in \cite{chen2019graph}. 
The main difference is that the GloRe unit reasons the relations within the same feature map via a standard graph, while the proposed BGR block reasons the cross relations between feature maps of different poses using a bipartite graph.

\subsection{Pose-to-Image Interaction and Aggregation}
As shown in Figure~\ref{fig:method}, the proposed IA block receives the appearance code $F_{t-1}^i$, the new source shape code $\tilde{F}_{t-1}^{pa}$, and the new target shape code $\tilde{F}_{t-1}^{pb}$ as inputs.
The IA block aims to simultaneously and interactively enhance $F_{t}^i$, $F_{t}^{pa}$ and $F_{t}^{pb}$.
Specifically, the shape codes are first concatenated and then  fed into two convolutional layers to produce the attention map $A_{t-1}$.
Mathematically,
\begin{equation}
\begin{aligned}
A_{t-1} = \sigma({\rm Conv}({\rm Concat}(\tilde{F}_{t-1}^{pa}, \tilde{F}_{t-1}^{pb}))),
\end{aligned}
\end{equation}
where $\sigma(\cdot)$ denotes the element-wise Sigmoid function. 

\hao{Appearance and shape features are crucial to generate the final person image since the appearance feature mainly focus on the texture and color information of clothes, and the shape feature mainly focus on the body orientation and size information. Thus, we propose the ``Appearance Code Enhancement'' to learn and enhance useful person appearance feature, while the ``Shape Code Enhancement'' to learn and enhance useful person shape feature.
Having both ``Appearance Code Enhancement'' and ``Shape Code Enhancement'' together can generate realistic person image.}

\noindent \textbf{Appearance Code Enhancement.} After obtaining $A_{t-1}$, the appearance $F_{t-1}^i$ is enhanced by:
\begin{equation}
\begin{aligned}
F_t^i = A_{t-1} \otimes  F_{t-1}^i  + F_{t-1}^i,
\end{aligned}
\label{eq:apperance}
\end{equation}
where $\otimes$ denotes the element-wise product. \hao{By multiplying with the attention map $A_{t-1}$, the new appearance code $F_t^i$ at certain locations can be either preserved or suppressed.}

\noindent \textbf{Shape Code Enhancement.}
\hao{
As the appearance code gets updated
through Eq. \eqref{eq:apperance}, the shape code should also be updated to synchronize the change, i.e., update where to sample and
put patches given the new appearance code. Therefore, the shape
code update should incorporate the new appearance code.}
Specifically, we concatenate $F_t^i $, $F_{t-1}^{pa}$ and $F_{t-1}^{pb}$, and pass them through two convolutional layers to obtain the updated shape codes $F_t^{pa}$ and $F_t^{pb}$ by splitting the result along the channel axis.
This process can be formulated as:
\begin{equation}
\begin{aligned}
F_{t}^{pa}, F_{t}^{pb} = {\rm Conv} ({\rm Concat}(F_t^i, \tilde{F}_{t-1}^{pa}, \tilde{F}_{t-1}^{pb})).
\end{aligned}
\end{equation}
\hao{In this way, both new shape codes $F_{t}^{pa}$ and $F_{t}^{pb}$ can synchronize the changes caused by the new appearance code $F_t^i$.}

\subsection{Attention-Based Image Fusion}
In the $T$-th IA block, we obtain the final appearance code $F_T^{i}$.
We then feed $F_T^{i}$ to an image decoder to generate the intermediate result $\tilde{I}_b$.
At the same time, we feed $F_T^{i}$ to an attention decoder to produce the attention mask $A_i$.

The attention encoder consists of several deconvolutional layers and a Sigmoid activation layer. Thus, the attention encoder aims to generate a one-channel attention mask $A_i$, in which each pixel value is between 0 to 1.
The attention mask $A_i$ aims to selectively pick useful content from both the input image $I_a$ and the intermediate result $\tilde{I}_b$ for generating the final result~$I_b^{'}$. This process can be expressed as:
\begin{equation}
\begin{aligned}
I_b^{'} = I_a \otimes  A_i + \tilde{I}_b \otimes (1 - A_i),
\end{aligned}
\label{eq:att}
\end{equation}
where $\otimes$ denotes an element-wise product.
In this way, both the image decoder and the attention decoder can interact with each other and ultimately produce better results.

\begin{figure*}[!t]
	\centering
	\includegraphics[width=1\linewidth]{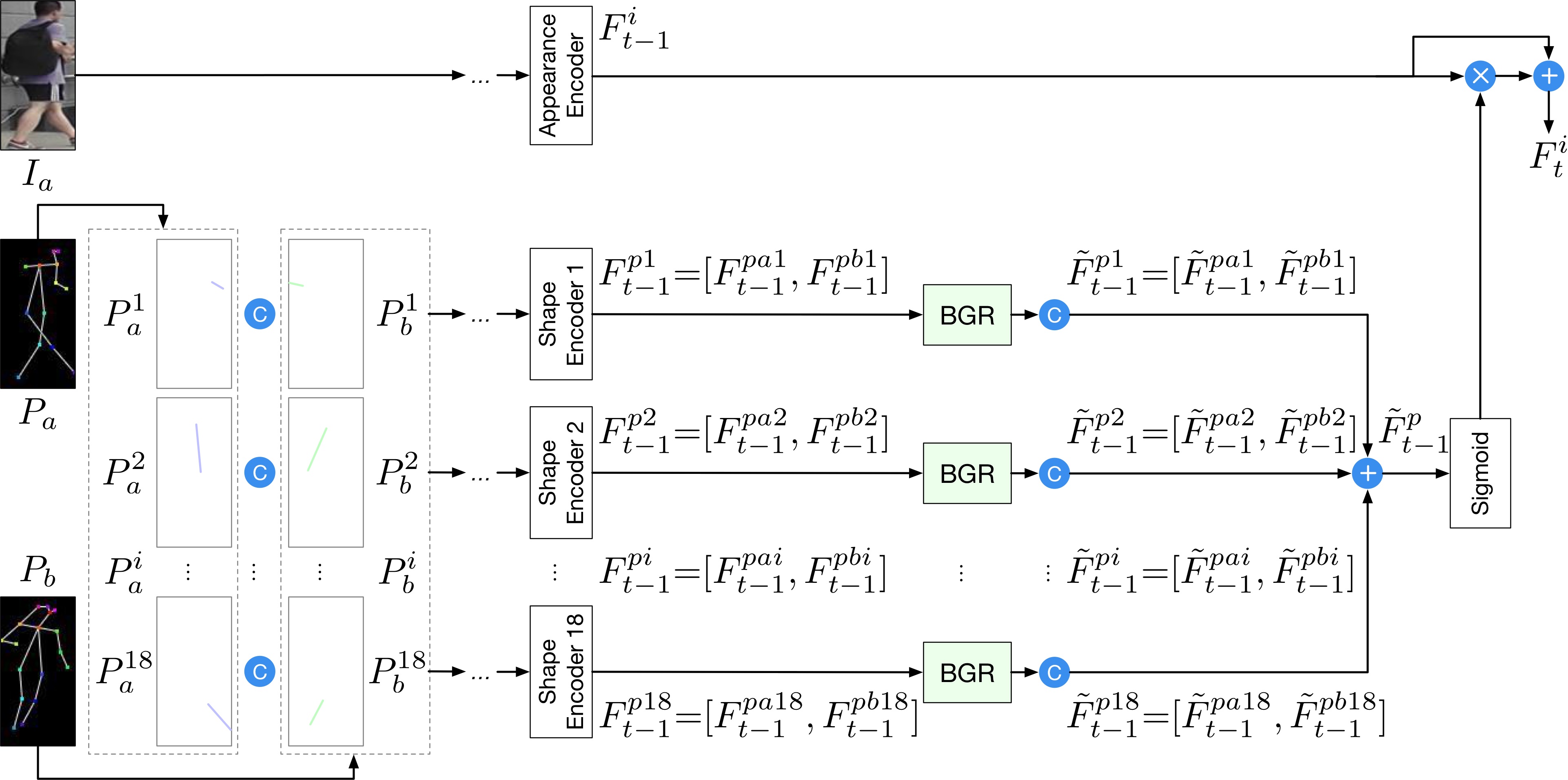}
	\caption{Illustration of the proposed PBGR block $t$, which consists of 18 branches. Each branch aims to model local transformations between each source sub-pose $F_{t-1}^{pai}$ and each target sub-pose $F_{t-1}^{pbi}$ in a bipartite graph via a BGR block presented in Figure~\ref{fig:blocks}.  
	\hao{Note that the shape encoders can share network parameters, so that no extra parameters are introduced, and the speed of training and testing is not significantly slow down.}
	}
	\label{fig:method_p}
\end{figure*}

\section{Part-Aware BiGraphGAN}
The proposed part-aware bipartite graph reasoning GAN (i.e., BiGraphGAN++) employs the same framework as BiGraphGAN, presented in Figure \ref{fig:method}, with the only difference being that we need to replace the BGR block from Figure \ref{fig:method} with the new PBGR from Figure~\ref{fig:method_p}.

\subsection{Part-Aware Bipartite Graph Reasoning}
The framework of the proposed PBGR block is shown in Figure \ref{fig:method_p}.
Specifically, we first follow OpenPose \cite{cao2017realtime} and decompose the overall source pose $P_a$ and target pose $P_b$ into 18 different sub-poses (i.e., $\{P_a^i\}_{i=1}^{18}$, and $\{P_b^i\}_{i=1}^{18}$) based on the inherent connection relationships among them.
Then the corresponding source and target sub-poses are concatenated and fed into the corresponding shape encoder to generate high-level feature representations.

Consider the $t$-th block given in Figure \ref{fig:method_p}. Each source and target sub-pose feature representation can be represented as $F_{t-1}^{pai}$ and $F_{t-1}^{pbi}$, respectively.
Then, the feature pair $[F_{t-1}^{pai}, F_{t-1}^{pbi}]$ is fed to the $i$-th BGR block to learn the local transformation for the $i$-th sub-pose, which can ease the learning and capture the change in pose of each part more precisely. 
Next, the updated feature representations $\tilde{F}_{t-1}^{pai}$ and $\tilde{F}_{t-1}^{pbi}$ are concatenated to represent the local transformation of the $i$-th sub-pose, i.e., $\tilde{F}_{t-1}^{pi}{=}[\tilde{F}_{t-1}^{pai}, \tilde{F}_{t-1}^{pbi}]$. 
Finally, we combine all the local transformations from all the different sub-poses to obtain the global transformation between the source pose $P_a$ and target pose $P_b$, which can be expressed as follows:
\begin{equation}
\begin{aligned}
\tilde{F}_{t-1}^{p} = \tilde{F}_{t-1}^{p1} + \tilde{F}_{t-1}^{p2} + \dots + \tilde{F}_{t-1}^{pi} + \dots + \tilde{F}_{t-1}^{p18} .
\end{aligned}
\label{eq:all}
\end{equation}

\subsection{Part-Aware Interaction and Aggregation}
The proposed part-aware IA block aims to simultaneously enhance $\tilde{F}_{t-1}^{p}$ and $F_{t-1}^{i} $.
Specifically, the pose feature  $\tilde{F}_{t-1}^{p}$ is fed into a Sigmoid activation layer to produce the attention map $A_{t-1}$.
Mathematically,
\begin{equation}
\begin{aligned}
A_{t-1} = \sigma(\tilde{F}_{t-1}^{p}),
\end{aligned}
\end{equation}
where $\sigma(\cdot)$ denotes the element-wise Sigmoid function. 
By doing so, $A_{t-1}$ provides important guidance for understanding the spatial deformation of each part region between the source and target poses, specifying which positions in the source pose should be sampled to generate the corresponding target pose. 

\noindent \textbf{Appearance Code Enhancement.} After obtaining $A_{t-1}$, the appearance $F_{t-1}^i$ is enhanced by:
\begin{equation}
\begin{aligned}
F_t^i = A_{t-1} \otimes  F_{t-1}^i  + F_{t-1}^i,
\end{aligned}
\end{equation}
where $\otimes$ denotes an element-wise product.

\noindent \textbf{Shape Code Enhancement.}
Next, we concatenate $F_t^i $ and $F_{t-1}^{pi}$, and pass them through two convolutional layers to obtain the updated shape codes $F_t^{pai}$ and $F_t^{pbi}$ by splitting the result along the channel axis.
This process can be formulated as:
\begin{equation}
\begin{aligned}
F_t^{pi} = & [F_{t}^{pai}, F_{t}^{pbi}] \\ = &  {\rm Conv} ({\rm Concat}(F_t^i, F_{t-1}^{pi})), i = 1, \cdots, 18.
\end{aligned}
\end{equation}
In this way, both new shape codes $F_{t}^{pai}$ and $F_{t}^{pbi}$ can synchronize the changes caused by the new appearance code $F_t^i$. 

\section{Model Training}
\noindent \textbf{Appearance and Shape Discriminators.} 
We adopt two discriminators for adversarial training.
Specifically, we feed the image-image pairs ($I_a$, $I_b$) and ($I_a$, $I_b^{'}$) into the appearance discriminator $D_{app}$ to ensure appearance consistency.
Meanwhile, we feed the pose-image pairs ($P_b$, $I_b$) and ($P_b$, $I_b^{'}$) into the shape discriminator $D_{sha}$ for shape consistency.
Both discriminators (i.e., $D_{app}$ and $D_{sha}$) and the proposed graph generator $G$ are trained in an end-to-end way, enabling them to enjoy mutual benefits from each other in a joint framework.

\noindent \textbf{Optimization Objectives.} We follow \cite{zhu2019progressive,tang2020xinggan} and use the adversarial loss $\mathcal{L}_{gan}$, the pixel-wise $L1$ loss $\mathcal{L}_{l1}$, and the perceptual loss $\mathcal{L}_{per}$ as our optimization objectives:
\begin{equation}
\begin{aligned}
\mathcal{L}_{full} = \lambda_{gan} \mathcal{L}_{gan} + \lambda_{l1} \mathcal{L}_{l1} + \lambda_{per} \mathcal{L}_{per},
\end{aligned}
\label{eq:loss}
\end{equation}
where $\lambda_{gan}$, $\lambda_{l1}$, and $\lambda_{per}$ control the relative importance of the three objectives.
For the perceptual loss, we follow \cite{zhu2019progressive,tang2020xinggan} and use the $Conv1\_2$ layer.

\noindent \textbf{Implementation Details.}
In our experiments, we follow previous work~\cite{zhu2019progressive,tang2020xinggan} and represent the source pose $P_a$ and the target pose $P_b$ as two 18-channel heat maps that encode the locations of 18 joints of a human body.
The Adam optimizer \cite{kingma2014adam} is employed to learn the proposed BiGraphGAN and BiGraphGAN++ for around 90K iterations with $\beta_1{=}0.5$ and $\beta_2{=}0.999$.

In our preliminary experiments, we found that as $T$ increases, the performance gets better and better. 
However, when $T$ reaches 9, the proposed models achieve the best results, and then the performance begins to decline. 
Thus, we set $T{=}9$ in the proposed graph generator.
Moreover, $\lambda_{gan}$, $\lambda_{l1}$, $\lambda_{per}$ in Equation~\eqref{eq:loss}, and the number of feature map channels $C$, are set to 5, 10, 10, and 128, respectively.
The proposed BiGraphGAN is implemented in PyTorch~\cite{paszke2019pytorch}. 

\section{Experiments}

\begin{table*}[!t]
	\centering
	\caption{Quantitative comparison of different methods on Market-1501 and DeepFashion for person pose generation. For all metrics, higher is better. ($\ast$) denotes the results tested on our testing set.}
	\begin{tabular}{rccccccc} \toprule
		\multirow{2}{*}{Method}  & \multicolumn{4}{c}{Market-1501} & \multicolumn{3}{c}{DeepFashion} \\ \cmidrule(lr){2-5} \cmidrule(lr){6-8} 
		& SSIM  $\uparrow$ & IS $\uparrow$  & Mask-SSIM $\uparrow$ & Mask-IS $\uparrow$ & SSIM  $\uparrow$ & IS  $\uparrow$ & PCKh  $\uparrow$   \\ \hline	
		PG2~\cite{ma2017pose}                                        & 0.253 & 3.460 & 0.792 & 3.435    & 0.762 & 3.090  & - \\
		DPIG~\cite{ma2018disentangled}                           & 0.099 & 3.483 & 0.614 & 3.491   & 0.614 & 3.228    & - \\
		Deform~\cite{siarohin2018deformable}                  & 0.290 & 3.185 & 0.805 & 3.502   & 0.756 & 3.439    & -\\ 
		C2GAN~\cite{tang2019cycle}                                & 0.282 & 3.349 & 0.811 & 3.510    & -     & -            & -\\ 
		BTF~\cite{albahar2019guided}                               & -     & -     & -     & -       & 0.767 & 3.220                   & -\\ \hline
		PG2$^\ast$~\cite{ma2017pose}                             & 0.261 & 3.495 & 0.782 & 3.367   & 0.773 & 3.163  & 0.89 \\ 
		Deform$^\ast$~\cite{siarohin2018deformable}       & 0.291 & 3.230 & 0.807 & 3.502   & 0.760 & 3.362 & 0.94 \\ 
		VUNet$^\ast$~\cite{esser2018variational}              & 0.266 & 2.965 & 0.793 & 3.549   & 0.763 & 3.440 & 0.93 \\
		PATN$^\ast$~\cite{zhu2019progressive}               & 0.311 & 3.323 & 0.811 & 3.773    & 0.773 & 3.209 & 0.96 \\  
		PoseStylizer$^\ast$~\cite{huang2020generating}  & 0.312 & 3.132 & 0.808 & 3.729    & 0.775 & 3.292 & 0.96\\
		XingGAN$^\ast$~\cite{tang2020xinggan}             & 0.313  & 3.506 & 0.816 & \textbf{3.872}  & 0.778 & 3.476 & 0.95\\ \hline
		BiGraphGAN (Ours)                                              & 0.325 &  3.329 & 0.818 & 3.695  & 0.778 & 3.430 & \textbf{0.97}  \\ 
		BiGraphGAN++ (Ours)                                           & \textbf{0.334}  &\textbf{3.592}   & \textbf{0.822} & 3.701 & \textbf{0.802} & \textbf{3.508} & \textbf{0.97} \\ \hline	
		Real Data                                                               & 1.000 & 3.890 & 1.000 & 3.706   & 1.000 & 4.053 & 1.00 \\	
		\bottomrule	
	\end{tabular}
	\label{tab:pose_reuslts}
\end{table*}

\begin{table*}[!ht]
	\centering
	\caption{Quantitative comparison of user study (\%) on Market-1501 and DeepFashion. `R2G' and `G2R' represent the percentage of real images rated as fake w.r.t.~all real images, and the percentage of generated images rated as real w.r.t. all generated images, respectively.}
	\begin{tabular}{rccccccc} \toprule
		\multirow{2}{*}{Method}  & \multicolumn{2}{c}{Market-1501} & \multicolumn{2}{c}{DeepFashion} \\ \cmidrule(lr){2-3} \cmidrule(lr){4-5} 
		& R2G $\uparrow$ & G2R $\uparrow$ & R2G $\uparrow$ & G2R $\uparrow$   \\ \hline	
		PG2~\cite{ma2017pose}                              & 11.20  & 5.50    & 9.20   & 14.90 \\
		Deform~\cite{siarohin2018deformable}    & 22.67 & 50.24  & 12.42 & 24.61 \\ 
		C2GAN~\cite{tang2019cycle}                      & 23.20 & 46.70  & -     & -     \\
		PATN~\cite{zhu2019progressive}   & 32.23 & 63.47  & 19.14 & 31.78 \\  
		BiGraphGAN (Ours)   & 35.76 & 65.91  & 22.39 & 34.16  \\	
		BiGraphGAN++ (Ours)  & \textbf{37.32} & \textbf{66.83} & \textbf{23.76} & \textbf{35.57} \\
		\bottomrule	
	\end{tabular}
	\label{tab:pose_ruser}
\end{table*}

\begin{figure*}[!ht]\small
	\centering
	\subfigure[]{\label{fig:market1}\includegraphics[width=0.81\linewidth]{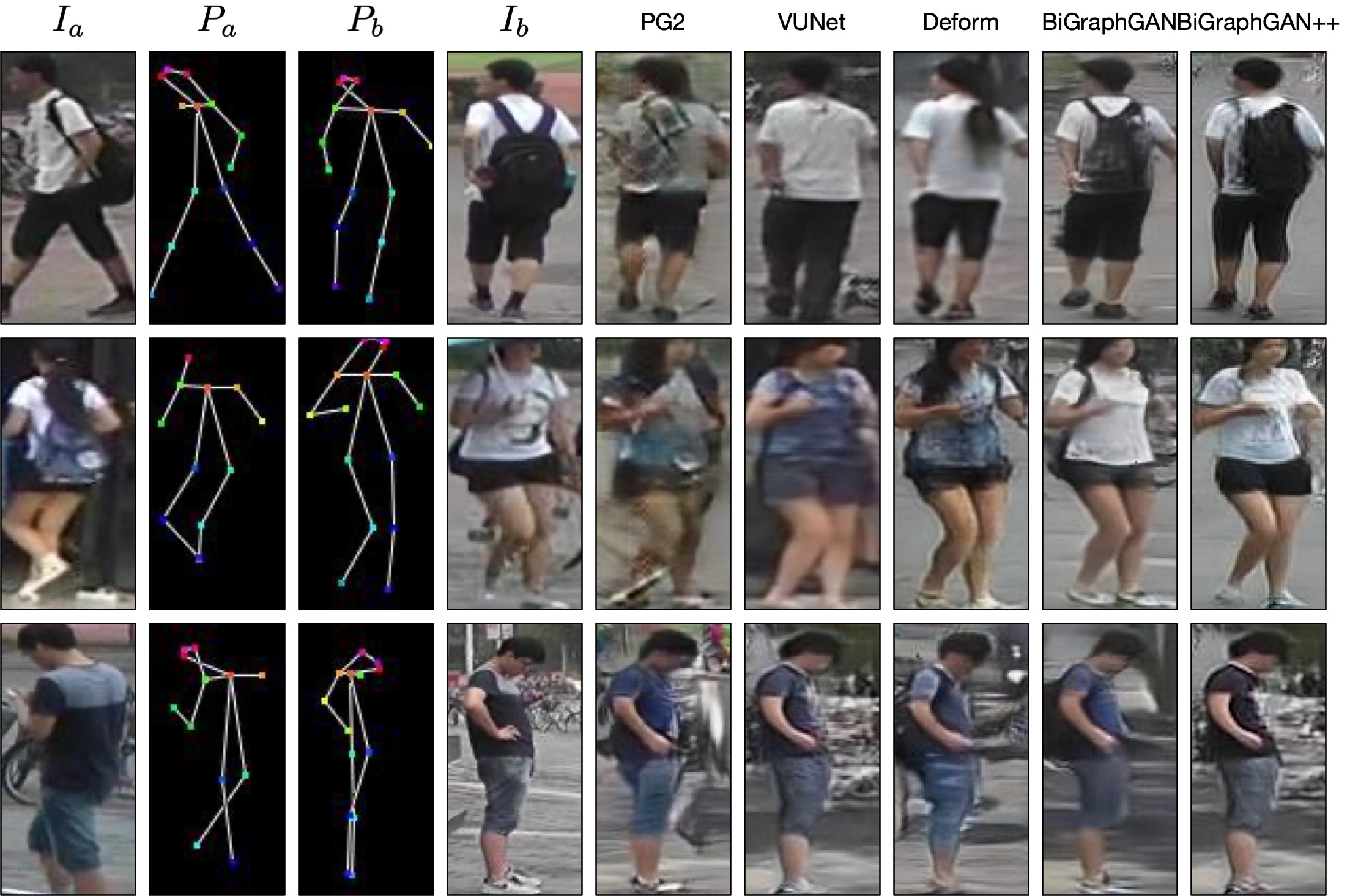}}
	\subfigure[]{\label{fig:market2}\includegraphics[width=0.81\linewidth]{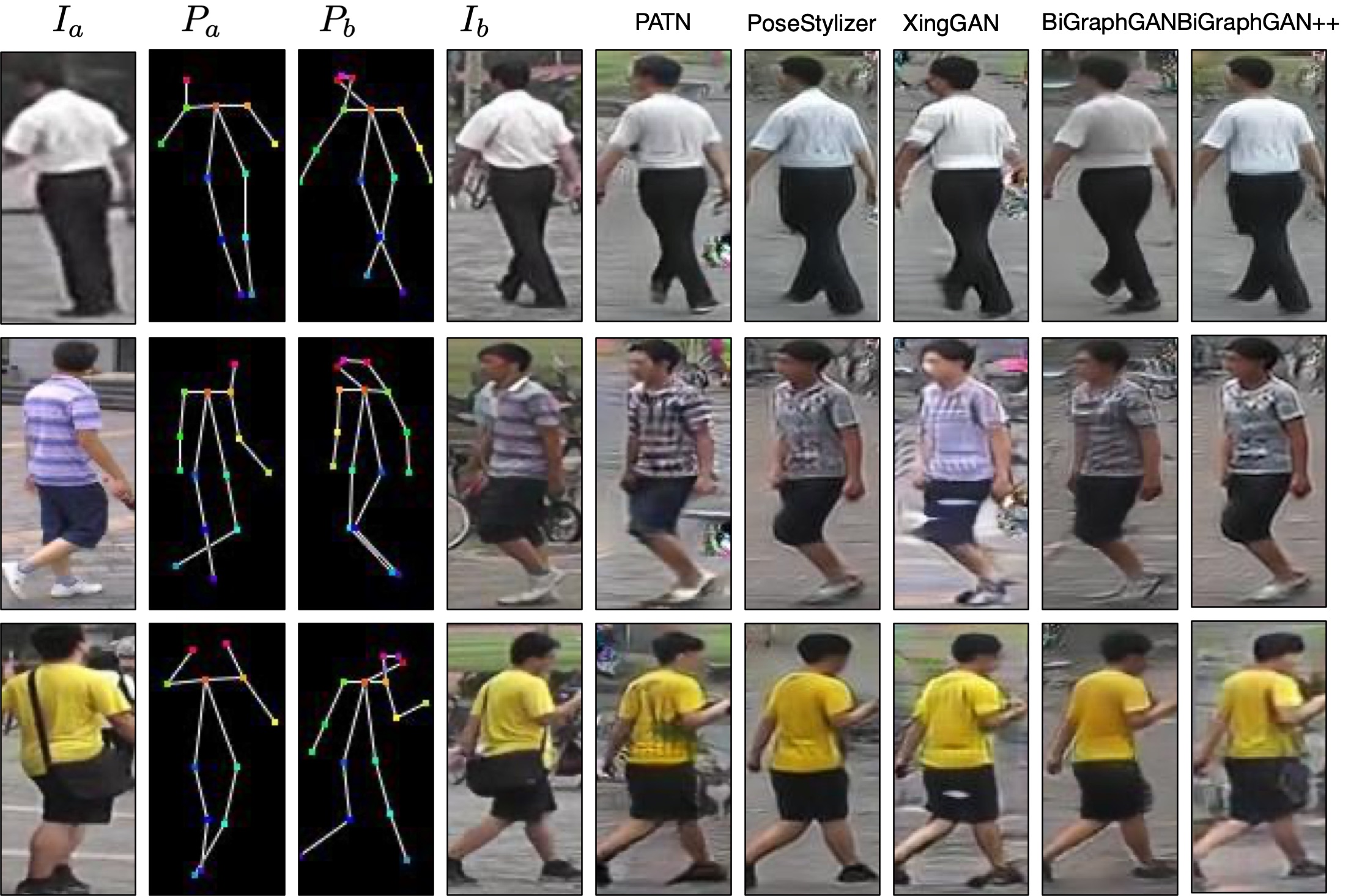}}
	\caption{Qualitative comparisons of person pose generation on Market-1501. (a) From left to right: Source Image ($I_a$), Source Pose ($P_a$), Target Pose ($P_b$), Target Image($I_b$), PG2~\cite{ma2017pose}, VUNet~\cite{esser2018variational}, Deform~\cite{siarohin2018deformable}, BiGraphGAN (Ours), and BiGraphGAN++ (Ours). (b) From left to right: Source Image ($I_a$), Source Pose ($P_a$), Target Pose ($P_b$), Target Image($I_b$), PATN~\cite{zhu2019progressive}, PoseStylizer~\cite{huang2020generating}, XingGAN \cite{tang2020xinggan}, BiGraphGAN (Ours), and BiGraphGAN++ (Ours).}
	\label{fig:mark_results}
\end{figure*}

\begin{figure*}[!htbp]\small
	\centering
	\subfigure[]{\label{fig:fashion1}\includegraphics[width=0.81\linewidth]{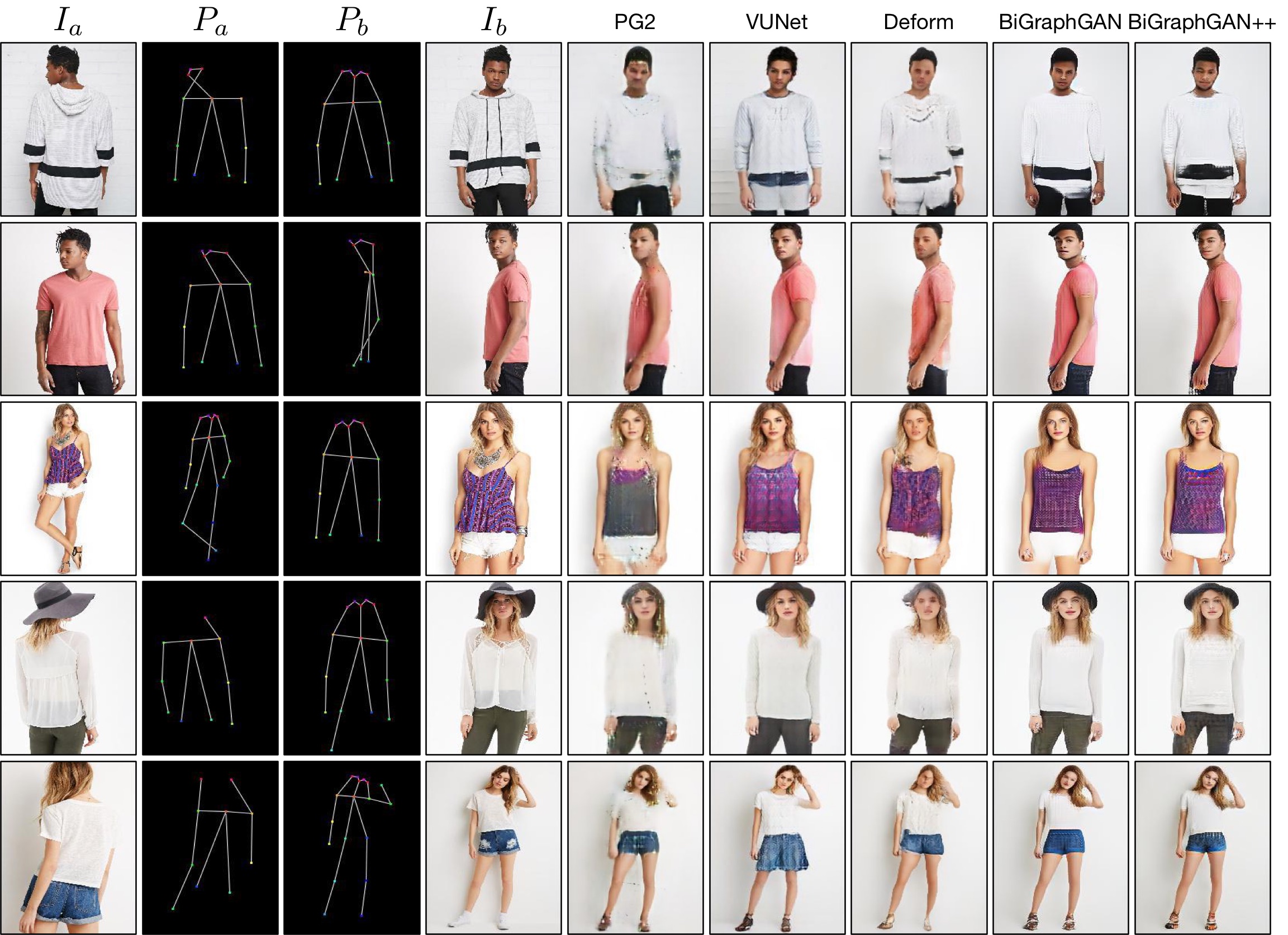}}
	\subfigure[]{\label{fig:fashion2}\includegraphics[width=0.72\linewidth]{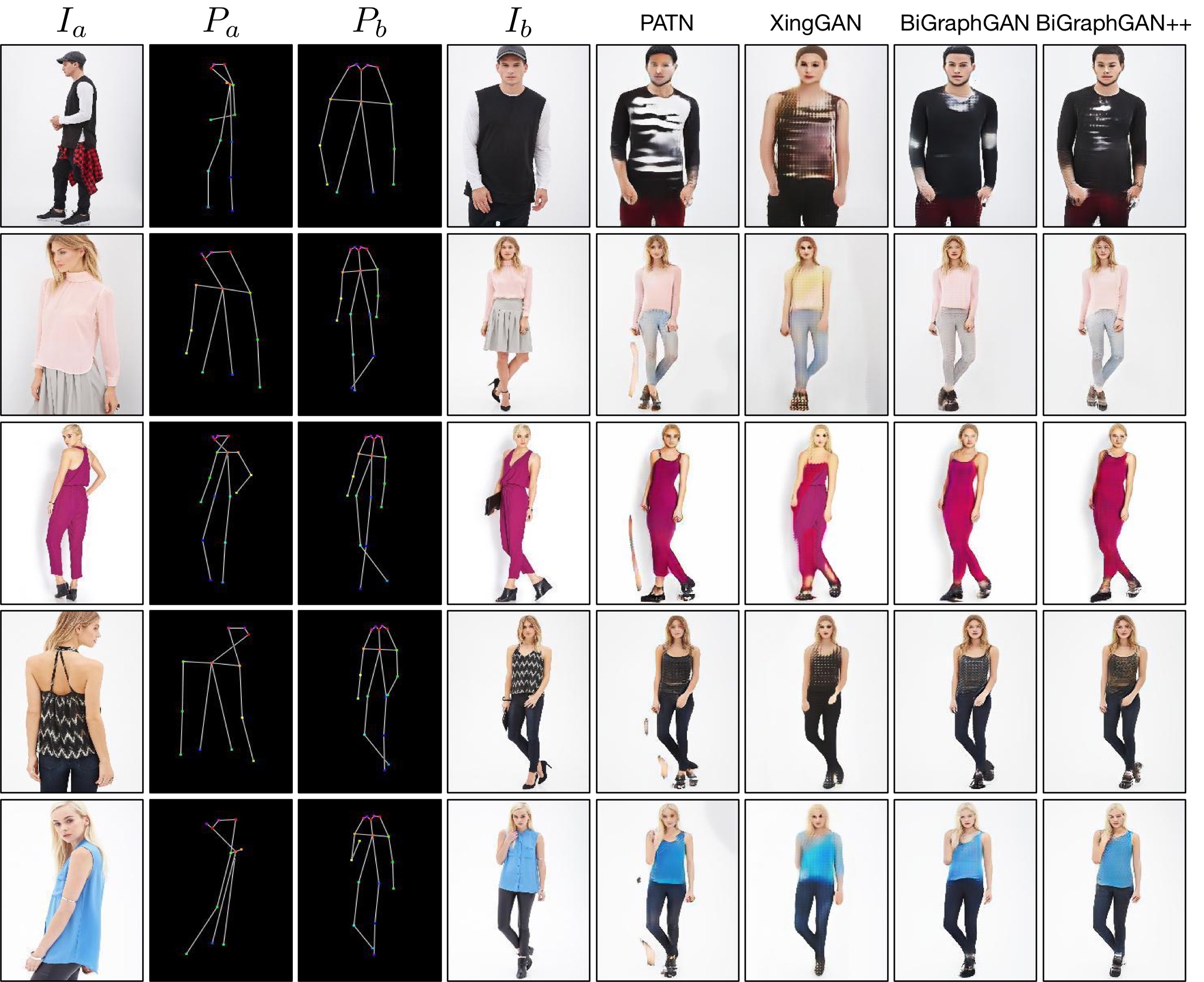}}
	\caption{Qualitative comparisons of person pose generation on DeepFashion. (a) From left to right: Source Image ($I_a$), Source Pose ($P_a$), Target Pose ($P_b$), Target Image($I_b$), PG2~\cite{ma2017pose}, VUNet~\cite{esser2018variational}, Deform~\cite{siarohin2018deformable}, BiGraphGAN (Ours), and BiGraphGAN++ (Ours). (b) From left to right: Source Image ($I_a$), Source Pose ($P_a$), Target Pose ($P_b$), Target Image($I_b$), PATN~\cite{zhu2019progressive}, XingGAN \cite{tang2020xinggan}, BiGraphGAN (Ours), and BiGraphGAN++ (Ours).}
	\label{fig:fashion_results}
\end{figure*}

\subsection{Person Pose Synthesis}

\noindent \textbf{Datasets.} 
We follow previous works \cite{ma2017pose,siarohin2018deformable,zhu2019progressive} and conduct extensive experiments on two public datasets, i.e., Market-1501 \cite{zheng2015scalable} and DeepFashion \cite{liu2016deepfashion}.
Specifically, we adopt the training/test split used in \cite{zhu2019progressive,tang2020xinggan} for fair comparison.
In addition, images are resized to $128 {\times} 64$ and $256 {\times} 256$ on Market-1501 and DeepFashion, respectively.
 
\noindent \textbf{Evaluation Metrics.}
We follow \cite{ma2017pose,siarohin2018deformable,zhu2019progressive} and employ Inception score (IS) \cite{salimans2016improved}, structural similarity index measure (SSIM) \cite{wang2004image}, and their masked versions (i.e., Mask-IS and Mask-SSIM) as our evaluation metrics to quantitatively measure the quality of the images generated by different approaches.
Moreover, we employ the percentage of correct keypoints (PCKh) score proposed in \cite{zhu2019progressive} to explicitly evaluate the shape consistency of the person images generated for the DeepFashion dataset.

\noindent \textbf{Quantitative Comparisons.}
We compare the proposed BiGraphGAN and BiGraphGAN++ with several leading person image synthesis methods, i.e., PG2~\cite{ma2017pose}, DPIG~\cite{ma2018disentangled}, Deform~\cite{siarohin2018deformable}, C2GAN~\cite{tang2019cycle}, BTF~\cite{albahar2019guided}, VUNet~\cite{esser2018variational}, PATN~\cite{zhu2019progressive}, PoseStylizer~\cite{huang2020generating}, and XingGAN~\cite{tang2020xinggan}.
Note that all of them use the same training data and data augmentation to train the models.

Quantitative comparison results are shown in Table~\ref{tab:pose_reuslts}. We observe that the proposed methods achieve the best results in most metrics, including SSIM and Mask-SSIM on Market-1501, and SSIM and PCKh on DeepFashion.
For other metrics, such as IS, the proposed methods still achieve better scores than the most related model, PATN, on both datasets.
These results validate the effectiveness of our proposed methods.

\noindent \textbf{Qualitative Comparisons.}
We also provide visual comparison results on both datasets in Figures~\ref{fig:mark_results} and~\ref{fig:fashion_results}.
As shown on the left of both figures, the proposed BiGraphGAN and BiGraphGAN++ generate remarkably better results than PG2~\cite{ma2017pose}, VUNet~\cite{esser2018variational}, and Deform~\cite{siarohin2018deformable} on both datasets.
To further evaluate the effectiveness of the proposed methods, we compare BiGraphGAN and BiGraphGAN++ with the most state-of-the-art models, i.e., PATN~\cite{zhu2019progressive}, PoseStylizer~\cite{huang2020generating}, and XingGAN~\cite{tang2020xinggan}, on the right of both figures.
We again observe that our proposed BiGraphGAN and BiGraphGAN++ generate clearer and more visually plausible person images than PATN, PoseStylizer, and XingGAN on both datasets.

\noindent \textbf{User Study.}
We also follow \cite{ma2017pose,siarohin2018deformable,zhu2019progressive} and conduct a user study to evaluate the quality of the generated images.
Specifically, we follow the evaluation protocol used in \cite{zhu2019progressive,tang2020xinggan} for fair comparison.
Comparison results of different methods are shown in Table~\ref{tab:pose_ruser}. We see that the proposed methods achieve the best results in all metrics, which further confirms that the images generated by the proposed BiGraphGAN and BiGraphGAN++ are more photorealistic.

\begin{figure*}[!t]
	\centering
	\includegraphics[width=1\linewidth]{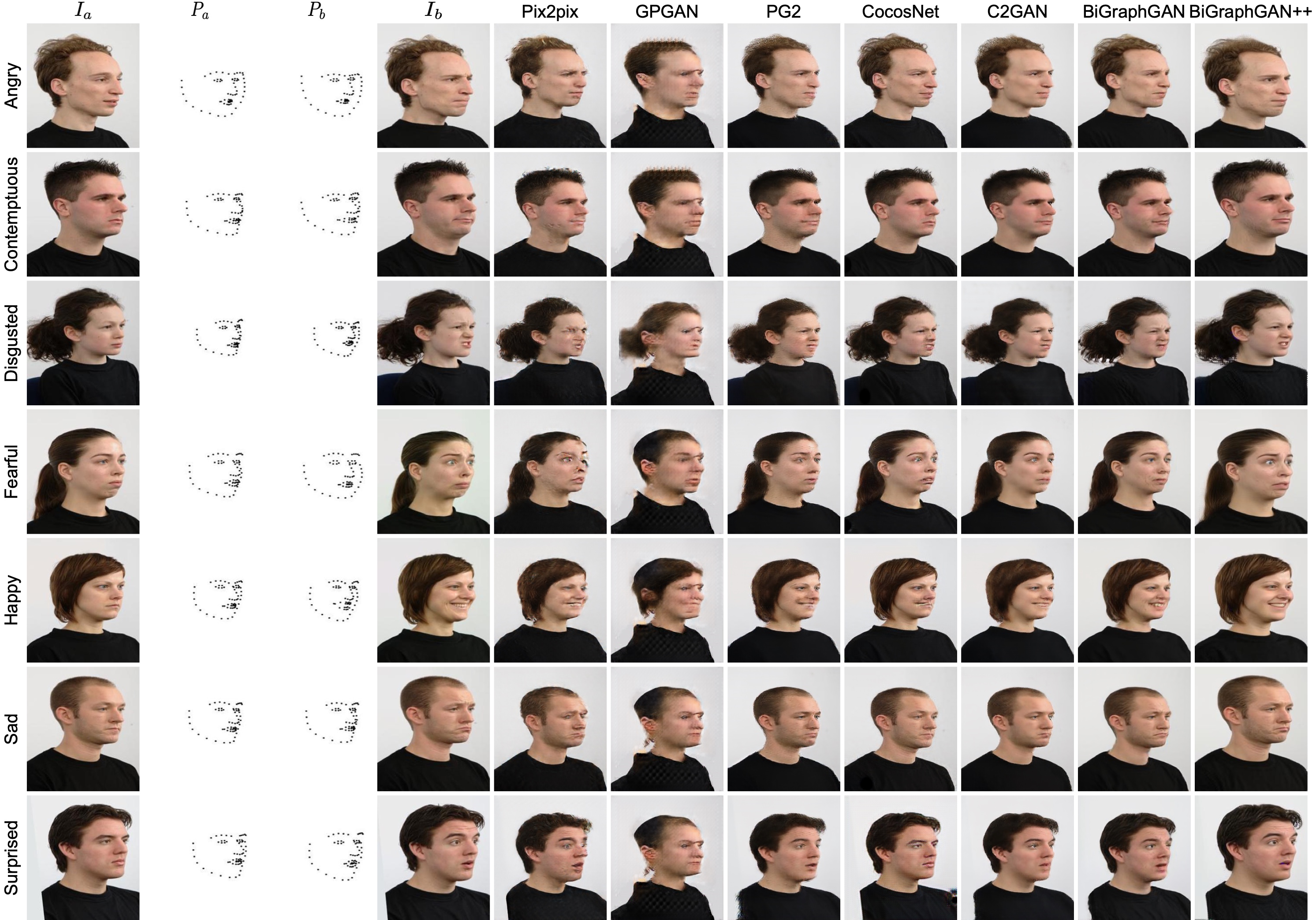}
	\caption{Qualitative comparisons of facial expression translation on Radboud Faces. From left to right: Source Image ($I_a$), Source Landmark ($P_a$), Target Landmark ($P_b$), Target Image ($I_b$),  Pix2pix~\cite{isola2017image}, GPGAN~\cite{di2018gp}, PG2~\cite{ma2017pose}, CocosNet \cite{zhang2020cross}, C2GAN \cite{tang2019cycle}, BiGraphGAN (Ours), and and BiGraphGAN++ (Ours).}
	\label{fig:face_results}
\end{figure*}

\begin{table*}[!ht]
	\centering
	\caption{Quantitative comparison of facial expression image synthesis on the Radboud Faces dataset. For all the metrics except LPIPS, higher is better.} 
	\begin{tabular}{rcccc} \toprule
		Method                          & AMT $\uparrow$  & SSIM $\uparrow$  & PSNR $\uparrow$   & LPIPS $\downarrow$ \\ \midrule
		Pix2pix~\cite{isola2017image}  & 13.4            & 0.8217           & 19.9971           & 0.1334 \\ 
		GPGAN~\cite{di2018gp}          & 0.3             & 0.8185           & 18.7211           & 0.2531 \\ 
		PG2~\cite{ma2017pose}          & 28.4            & 0.8462           & 20.1462           & 0.1130 \\  
		CocosNet \cite{zhang2020cross} & 31.3 & 0.8524 & 20.7915 & 0.0985 \\
		C2GAN \cite{tang2019cycle}  & 34.2          & 0.8618           & 21.9192           & 0.0934 \\   
		BiGraphGAN (Ours)               & 37.9 &  0.8644 & 27.5923 & 0.0806 \\ 
		BiGraphGAN++ (Ours)           & \textbf{39.1} & \textbf{0.8665} & \textbf{29.3917} & \textbf{0.0798} \\ \bottomrule	
	\end{tabular}
	\label{tab:result_face}
\end{table*}

\subsection{Facial Expression Synthesis}
\noindent\textbf{Datasets.} 
The Radboud Faces dataset~\cite{langner2010presentation} is used to conduct experiments on the facial expression generation task.
This dataset consists of over 8,000 face images with eight different facial expressions, i.e., neutral, angry, contemptuous, disgusted, fearful, happy, sad, and surprised.

We follow C2GAN \cite{tang2019cycle} and select 67\% of the images for training, while the remaining 33\% are used for testing. 
We use the public software OpenFace~\cite{amos2016openface} to extract facial landmarks.
For the facial expression-to-expression translation task, we combine two different facial expression images of the same person to form an image pair for training (e.g., neutral and angry). 
Thus, we obtain 5,628 and 1,407 image pairs for the training and testing sets, respectively.

\noindent\textbf{Evaluation Metrics.} 
We follow C2GAN \cite{tang2019cycle} and first adopt SSIM~\cite{wang2004image}, peak signal-to-noise ratio
(PSNR), and learned perceptual image patch similarity (LPIPS) \cite{zhang2018unreasonable} for quantitative evaluation.
Note that both SSIM and PSNR measure the image quality at a pixel level, while LPIPS evaluates the generated images at a deep feature level. 
Next, we again follow C2GAN and adopt the amazon mechanical turk (AMT) user study to evaluate the generated facial images.

\noindent \textbf{Quantitative Comparisons.}
To evaluate the effectiveness of the proposed BiGraphGAN, we compare it with several leading facial image generation methods, i.e., Pix2pix~\cite{isola2017image}, GPGAN~\cite{di2018gp}, PG2~\cite{ma2017pose}, CocosNet \cite{zhang2020cross}, and C2GAN \cite{tang2019cycle}.

The results in terms of SSIM, PSNR, and LPIPS are shown in Table \ref{tab:result_face}.
We observe that the proposed BiGraphGAN  and BiGraphGAN++ achieve the best scores in all three evaluation metrics, confirming the effectiveness of our methods.
Notably, the proposed BiGraphGAN is 5.6731 points higher than the current best method (i.e., C2GAN) in the PSNR metric.

\noindent \textbf{Qualitative Comparisons.}
We also provide qualitative results compared with the current leading models in Figure~\ref{fig:face_results}.
We observe that GPGAN performs the worst among all comparison models.
Pix2pix can generate correct expressions, but the faces are distorted.
Moreover, the results of PG2 tend to be blurry. Compared with these methods, the results generated by the proposed BiGraphGAN are smoother, sharper, and contain more convincing details. 

\begin{figure*}[t]
	\centering
	\subfigure[]{\label{fig:ablation1}\includegraphics[width=0.551\linewidth]{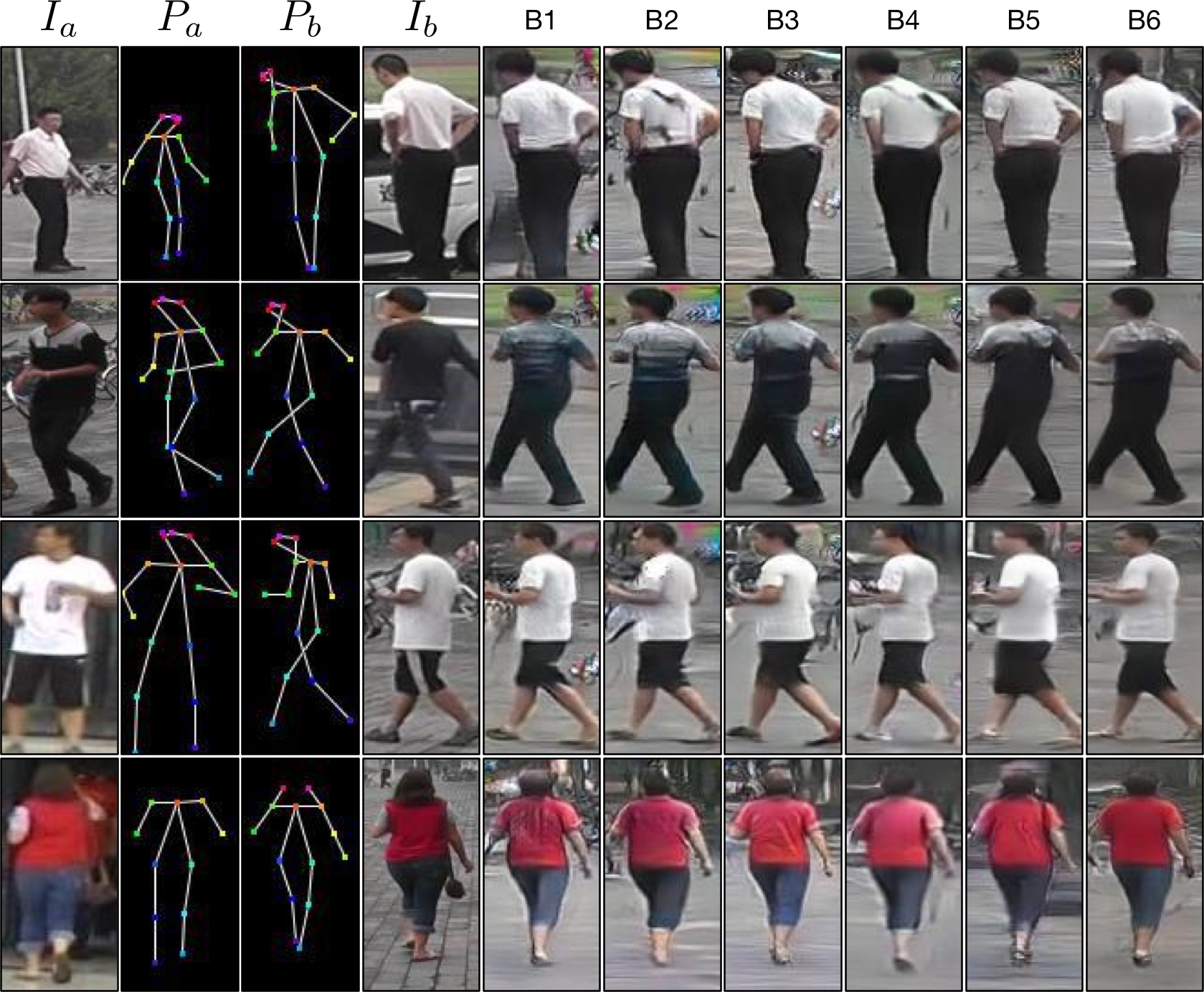}}
	\subfigure[]{\label{fig:ablation2}\includegraphics[width=0.44\linewidth]{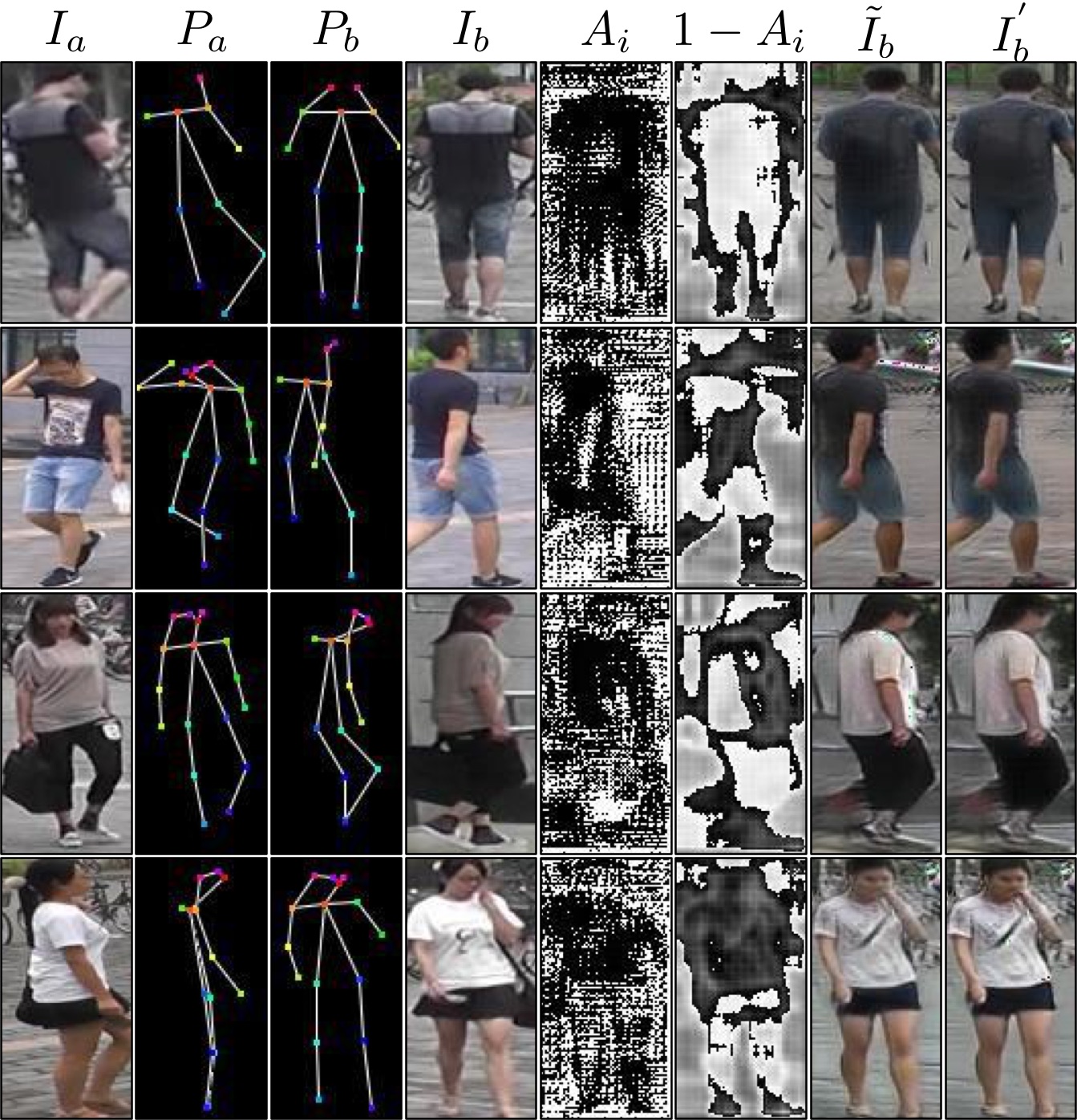}}
	\caption{Qualitative comparison of ablation study on Market-1501. (a) Qualitative comparisons of different baselines of the proposed BiGraphGAN. (b) Visualization of the learned attention masks and intermediate results.}
	\label{fig:ablation}
\end{figure*}

\begin{table*}[!t]
	\centering
	\caption{Ablation study of the proposed BiGraphGAN on Market-1501 for person pose generation. For both metrics, higher is better.}
	\begin{tabular}{lcc} \toprule
		Baselines of BiGraphGAN & SSIM $\uparrow$ & Mask-SSIM $\uparrow$    \\ \midrule
		B1: Our Baseline & 0.305 & 0.804 \\ 
		B2: B1 + B2A     & 0.310 & 0.809 \\
		B3: B1 + A2B     & 0.310 & 0.808 \\
		B4: B1 + A2B + B2A (Sharing) & 0.322 & 0.813 \\
		B5: B1 + A2B + B2A (Non-Sharing) & 0.324 & 0.813 \\
		B6: B5 + AIF & \textbf{0.325} & \textbf{0.818} \\
		\bottomrule	
	\end{tabular}
	\label{tab:ablation}
\end{table*}

\noindent \textbf{User Study.}
Following C2GAN~\cite{tang2019cycle}, we conduct a user study to evaluate the quality of the images generated by different models, i.e., Pix2pix~\cite{isola2017image}, GPGAN~\cite{di2018gp} , PG2~\cite{ma2017pose}, CocosNet \cite{zhang2020cross}, and C2GAN \cite{tang2019cycle}.
Comparison results are shown in Table~\ref{tab:result_face}.
We observe that the proposed BiGraphGAN achieves the best results, which further validates that the images generated by the proposed model are more photorealistic.

\subsection{Ablation Study}
We perform extensive ablation studies to validate the effectiveness of each component of the proposed BiGraphGAN on the Market-1501 dataset. 

\noindent \textbf{Baselines of BiGraphGAN.}
The proposed BiGraphGAN has six baselines (i.e., B1, B2, B3, B4, B5, B6), as shown in Table \ref{tab:ablation} and Figure~\ref{fig:ablation} (left).
B1 is our baseline.  
B2 uses the proposed B2A branch to model the cross relations from the target pose to the source pose.
B3 adopts the proposed A2B branch to model the cross relations from the source pose to the target pose.
B4 combines both the A2B and B2A branches to model the cross relations between the source pose and the target pose.
Note that both GCNs in B4 share parameters.
B5 employs a non-sharing strategy between the two GCNs to model the cross relations.
B6 is our full model and employs the proposed AIF module to enable the graph generator to attentively determine which part is most useful for generating the final person image.

\noindent \textbf{Ablation Analysis.}
The results of the ablation study are shown in Table \ref{tab:ablation} and Figure~\ref{fig:ablation} (left).
We observe that both B2 and B3 achieve significantly better results than B1, proving our initial hypothesis that modeling the cross relations between the source and target pose in a bipartite graph will boost the generation performance.
In addition, we see that B4 outperforms B2 and B3, demonstrating the effectiveness of modeling the symmetric relations between the source and target poses.
B5 achieves better results than B4, which indicates that using two separate GCNs to model the symmetric relations will improve the generation performance in the joint network.
B6 is better than B5, which clearly proves the effectiveness of the proposed attention-based image fusion strategy.

Moreover, we show several examples of the learned attention masks and intermediate results in Figure~\ref{fig:ablation} (right)
We can see that the proposed module attentively selects useful content from both the input image and intermediate result to generate the final result, thus validating our design motivation.

\noindent \textbf{BiGraphGAN vs. BiGraphGAN++.}
We also provide comparison results of BiGraphGAN and BiGraphGAN++ on both Market-1501 and DeepFashion. 
The results for person pose image generation are shown in Tables \ref{tab:pose_reuslts} and \ref{tab:pose_ruser}.
We see that BiGraphGAN++ achieves much better results in most metrics, indicating that the proposed PBGR module does indeed learn the local transformations among body parts, thus improving the generation performance.
From the visualization results in Figures~\ref{fig:mark_results} and \ref{fig:fashion_results}, we can see that BiGraphGAN++ generates more photorealistic images with fewer visual artifacts than BiGraphGAN, on both datasets.
The same conclusion can be drawn from the facial expression synthesis task, as shown in Table \ref{tab:result_face} and Figure \ref{fig:face_results}. Overall, the proposed BiGraphGAN++ can achieve better reuslts than BiGraphGAN on both challenging tasks, validating the effectiveness of our network design.

\section{Conclusion}
In this paper, we propose a novel bipartite graph reasoning GAN (BiGraphGAN) framework for both the challenging person pose and facial image generation tasks.
We introduce two novel blocks, i.e., the bipartite graph reasoning (BGR) block and interaction-and-aggregation (IA) block.
The former is employed to model the long-range cross relations between the source pose and the target pose in a bipartite graph.
The latter is used to interactively enhance both a person's shape and appearance features.

To further capture the detailed local structure transformations among body parts, we propose a novel part-aware bipartite graph reasoning (PBGR) block.
Extensive experiments in terms of both human judgments and automatic evaluation demonstrate that the proposed BiGraphGAN achieves remarkably better performance than the state-of-the-art approaches on three challenging datasets.
Lastly, we believe that the proposed method will inspire researchers to explore the cross-contextual information in other vision task.


\small
\bibliographystyle{plain}
\bibliography{ref}

\end{document}